\documentclass{article}

\usepackage{microtype}
\usepackage{graphicx}
\usepackage{subcaption}
\usepackage{booktabs} 

\usepackage{hyperref}
\usepackage{tabularx}
\usepackage{multirow}

\usepackage[ruled,vlined]{algorithm2e}

\usepackage[accepted]{icml2025}

\usepackage{amsmath}
\usepackage{amssymb}
\usepackage{mathtools}
\usepackage{amsthm}
\usepackage{mdframed}
\mdfsetup{%
linecolor=white,
backgroundcolor=gray!20,
}
\usepackage[capitalize,noabbrev]{cleveref}

\theoremstyle{plain}

\theoremstyle{definition}

\theoremstyle{remark}

\usepackage[textsize=tiny]{todonotes}

\icmltitlerunning{Accepted in ICML 2025 Workshop on
Machine Learning for Wireless Communication and Networks (ML4Wireless)}

\begin{document}

\twocolumn[
\icmltitle{Hierarchical and Collaborative LLM-Based Control for Multi-UAV Motion and Communication in Integrated Terrestrial and Non-Terrestrial Networks}



\icmlsetsymbol{equal}{*}

\begin{icmlauthorlist}
\icmlauthor{Zijiang Yan}{equal,yorku}
\icmlauthor{Hao Zhou}{equal,McGill}
\icmlauthor{Jianhua Pei}{hust}
\icmlauthor{Hina Tabassum}{yorku}

\end{icmlauthorlist}

\icmlaffiliation{yorku}{Department of Electrical Engineering and Computer Science, York University, Toronto, Canada}
\icmlaffiliation{McGill}{Department of Computer Science, McGill University, Montréal, Canada}
\icmlaffiliation{hust}{School of Electrical and Electronic Engineering, Huazhong University of Science and Technology, Wuhan, China}

\icmlcorrespondingauthor{Hao Zhou}{hao.zhou4@mail.mcgill.ca}

\icmlkeywords{Machine Learning, ICML}

\vskip 0.3in
]
\fancyhead{}
\fancyhead[C]{ Accepted in ICML 2025 Workshop on Machine Learning for Wireless Communication and Networks (ML4Wireless)}
\thispagestyle{fancy}
\renewcommand{\headrulewidth}{0pt}      
\pagestyle{plain} 
\printAffiliationsAndNotice{\icmlEqualContribution}

\begin{abstract}
Unmanned aerial vehicles (UAVs) have been widely adopted in various real-world applications. However, the control and optimization of multi-UAV systems remain a significant challenge, particularly in dynamic and constrained environments.
This work explores the joint motion and communication control of multiple UAVs operating within integrated terrestrial and non-terrestrial networks that include high-altitude platform stations (HAPS). Specifically, we consider an aerial highway scenario in which UAVs must accelerate, decelerate, and change lanes to avoid collisions and maintain overall traffic flow. Different from existing studies, we propose a novel hierarchical and collaborative method based on large language models (LLMs). In our approach, an LLM deployed on the HAPS performs UAV access control, while another LLM onboard each UAV handles motion planning and control. This LLM-based framework leverages the rich knowledge embedded in pre-trained models to enable both high-level strategic planning and low-level tactical decisions. This knowledge-driven paradigm holds great potential for the development of next-generation 3D aerial highway systems. Experimental results demonstrate that our proposed collaborative LLM-based method achieves higher system rewards, lower operational costs, and significantly reduced UAV collision rates compared to baseline approaches.
\end{abstract}

\section{Introduction}
Unmanned aerial vehicles (UAVs) have seen widespread adoption across numerous application areas due to the ease of deployment, steadily reduced production costs, and the availability of line-of-sight (LOS) channels~\cite{yu2022deep}. 
Depending on the mission, a UAV may act as user equipment requiring cellular access or operate as an aerial base station (BS) to extend network coverage. Nonetheless, for UAVs flying beyond visual line-of-sight, robust command-and-control links are indispensable to guarantee both the operation reliability and security.
Meanwhile, High Altitude Platform Stations (HAPS) has recently emerged as a crucial part of 6G Non-Terrestrial Networks (NTNs). HAPS, known for its extensive propagation coverage, long-distance reach, and utilization of eco-friendly energy sources like solar and wind power, is progressively becoming a fundamental component of 6G networks~\cite{kurt2021vision}.

Compared to the control of an individual UAV, the joint coordination of multiple UAVs inherently incurs substantially greater complexity, since it must address not only the increased state-space dimensionality but also the inter-UAV synchronization required to maintain formation and collaborative mission objectives~\cite{cherif20213d, kotarski2020design}.
In addition, when these UAV swarms are integrated with a HAPS within a NTN, additional challenges may arise from the heterogeneity of air-to-air and air-to-ground links. 
Resource allocation, network topology management, and control-plane  algorithms will jointly contribute to the overall complexity across multi-UAV control, further complicating system design, real-time decision making, and end-to-end quality-of-service (QoS) guarantees.

Existing research primarily focuses on optimizing cellular link availability and QoS using reinforcement learning (RL) algorithms, without considering multi-UAV aerial traffic flow and motion dynamics of UAVs. 
In \cite{cherif2022cellular}, the authors proposed a RL algorithm that considers disconnectivity, handovers, and energy consumption for trajectory planning and cell association in cargo UAVs. 
Nesrine \textit{et al.} leverage RL and propose an algorithm for joint trajectory planning and cell association in \cite{cherif2023rl}, to minimize energy consumption and number of handoff events with QoS constraints. 
The authors in \cite{chen2020efficient} present strategies based on deep learning to predict handovers in mmWave communications and optimize handover (HO) rates and radio link quality for known UAV trajectories. 
However, most existing algorithm's actions only consider the direction of motion with no speed and lane considerations\cite{li2020novel}. These works have not considered the motion dynamics factors, such as acceleration, deceleration, and lane changes of multi-UAVs~\cite{zhang2022game}.  
For instance, collision avoidance is crucial for guaranteeing the performance for multi-UAV traffic flows. 
Specifically, in densely populated aerial corridors, each UAV must continuously monitor the positions and velocities of its neighbors and execute timely trajectory adjustments to preserve safe separation distances. Failure to do so can lead to emergency maneuvers that disrupt coordinated flight patterns, introduce significant latency in UAV task execution.

Different from existing studies, this work considers a more challenging scenario of a 3D aerial highway, in which multiple UAVs have to jointly share the highway lanes by acceleration/deceleration, lane changes, etc.  
Meanwhile, from the communication perspective, the UAVs are expected to achieve higher data rates and minimal handover losses. 
The transportation decisions are closely coupled with the wireless communication system performance. 
For instance, increasing speed can increase transportation traffic flow, but it will also lead to frequent handovers, which can negatively impact the UAV communications.

To better handle such complexity, this work employs a hierarchical and collaborative control method based on large language models (LLMs). Leveraging the intrinsic strengths of LLMs, our approach harnesses their extensive pre-trained knowledge base to support both high-level planning and fine-grained decision-making. Through in-context learning, the LLM dynamically interprets real-time traffic patterns, channel state information, and mission objectives, allowing it to infer optimal acceleration, deceleration, and lane-change policies without requiring exhaustive task-specific retraining.
In addition, the instruction-following capabilities further enable seamless incorporation of safety constraints, such as minimum inter-UAV spacing and maximum allowable handover rates, directly into the decision pipeline, ensuring both transportation and communication objectives are satisfied.

Existing studies have explored various application scenarios of LLM-enabled 6G networks, including grounding and alignment~\cite{xu2024large}, prompt engineering~\cite{zhou2025large}, network design~\cite{qiu2024large}, power control~\cite{zhou2024large}, etc.  However, this work is different by including hierarchical and collaborative multiple LLM interactions.  
In particular, we consider one LLM deployed in the HAPS to manage the UAV-HAPS connections, and another LLM for UAV action control such as acceleration, deceleration, and lane change.
By unifying transportation and communication control within an LLM‐driven architecture, our method can significantly reduce the design complexity associated with traditional model‐based controllers. 
It also accelerates adaptation to novel operational scenarios, such as varying UAV densities or spectrum availability, through prompt and instruction‐guided updates. 
This end‐to‐end, knowledge‐rich paradigm holds significant promise for next‐generation 3D aerial highways, where the seamless interplay of mobility and connectivity is paramount.
Finally, the simulations show that our proposed technique can achieve higher rewards, lower transportation and communication costs, and lower collision rates for multi-UAV scenarios.

\section{System Model}
\label{systemmodel}

\begin{figure*}[t]
\label{fig:uav-haps-system-model}
\centering
\begin{tabular}{lccccc}
\includegraphics[width=1\linewidth]{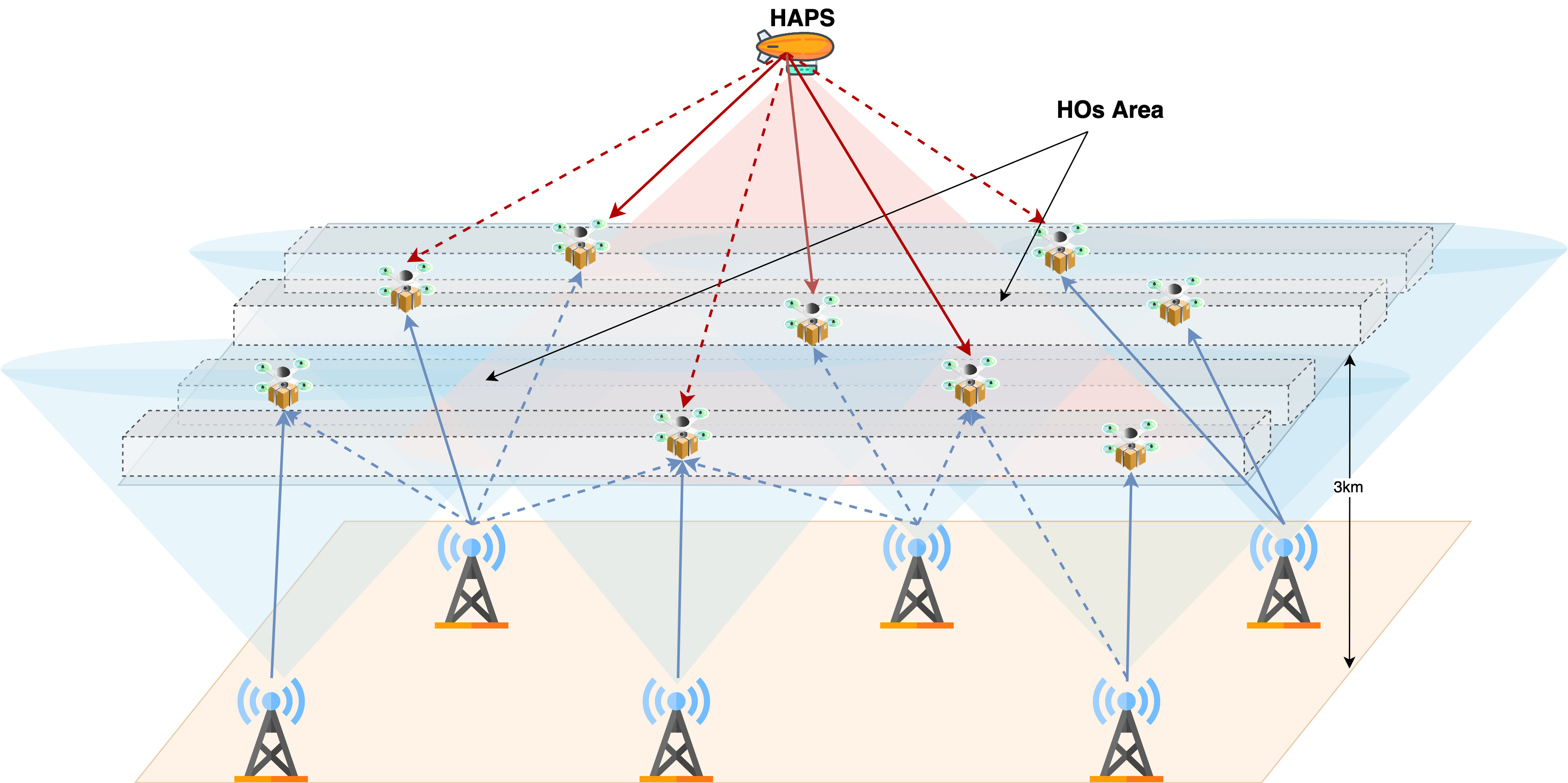}\hspace{-1cm}&\hspace{0.25cm}
\end{tabular}
\vspace{-2mm}
\caption {Illustration of the proposed aerial network model. }
\label{fig:bdq_agent}
\end{figure*}

Fig. \ref{fig:bdq_agent} shows our considered aerial highway scenario, where $N_{\mathrm{UAV}}$ UAVs are flying along the highway lanes, and UAVs can connect HAPS or terrestrial BSs.
The HAPS and terrestrial BSs form a set $\mathcal{B}=\{b_1,\ldots, b_{N_R}, b_{\mathrm{HAPS}}\}$, where each BS is denoted by $i$ for $\forall i \in \mathcal{B}$.
To simulate the UAVs’ movements on a given aerial highway, we consider the continuous intelligent driver model that models acceleration as in \cite{Treiber2013}.

\subsection{G2A Channel Model}

Firstly, the ground-to-air (G2A) channel model mainly depends on the BS's antenna gain, and the experienced path loss, and line-of-sight (LoS) probability \cite{3gpp777}.

In cellular‐connected aerial systems, UAVs depend on the sidelobe emissions of BS antennas to establish links with terrestrial infrastructure. Consequently, an accurate representation of BS 3D radiation characteristics is essential for reliable UAV connectivity. 
We adopt the 3GPP antenna pattern specification \cite{3gpp777}, which closely approximates real‐world radiation behavior. Conforming to this model, each BS is partitioned into three sectors, each armed with cross‐polarized elements arranged in a uniform linear array (ULA).
Every antenna element delivers a peak gain of $B_{\max}=8$ dBi along its principal lobe \cite{3gpp777}, while sidelobe gains vary according to the UAV’s azimuth and elevation angles relative to the array \cite{cherif2022cellular}:
\begin{equation}
    B_{\mathrm{az}}(\phi^{ij}_t) =\min \left\{ 12 \left( {\phi^{ij}_t}/{\phi_\mathrm{3dB}} \right),\mathrm{B_m}  \right\},
\end{equation}
\begin{equation}
    B_{\mathrm{el}}(\zeta^{ij}_t) =\min \left\{ 12 \left( {\zeta^{ij}_t}/{\zeta_\mathrm{3dB}} \right), \mathrm{SLA}  \right\},
\end{equation}
where $\phi^{ij}_t$ and $\zeta^{ij}_t$ are the azimuth and elevation angles between BS $i$ and UAV $j$, and $\phi_\mathrm{3dB}=\zeta_\mathrm{3dB}=\frac{65\pi}{180}$ at 3dB bandwidths. $\mathrm{B_m} $ and $\mathrm{SLA}$ represent the antenna nulls thresholds, respectively. 
The, the gain of antenna element is
{\small
\begin{equation}
        B(\zeta^{ij}_t, \phi^{ij}_t) = B_{\mathrm{max}} 
    - \min \{- (B_{\mathrm{az}}(\phi^{ij}_t) + B_{\mathrm{el}}(\zeta^{ij}_t) ) , B_m \}.
\end{equation}}
Assuming that one BS equipped with $N$ antennas, and the array factor of the ULA of BS $i$ and UAV $j$ is
\begin{equation}
    \mathrm{F}(\zeta^{ij}_t) = \frac{\sin(\frac{N\pi}{2} (\sin \zeta^{ij}_t - \sin \zeta_i^d))}{\sqrt{N} \sin(\frac{\pi}{2} (\sin\zeta^{ij}_t - \sin\zeta_i^d))},
\end{equation}
where $\zeta_i^d$ is the down-tilt of BS $i$'s ULA. Finally, the array radiation pattern from BS $i$ to UAV $j$ is
\begin{equation}
B^{ij}_t = B(\zeta^{ij}_t,\phi^{ij}_t) + \mathrm{F}(\zeta^{ij}_t).
\end{equation}

Then, the probability of UAV $j$ has a LoS connection with BS $i$ depends on the altitude of the UAV and the surrounding environment, and the LoS probability is given by 
\begin{equation}
    P_{\mathrm{LoS}}(q^{ij}_t)= 
\begin{cases}
1,\quad &d^{ij}_t \leq d_1 \\
{d_1}/{d^{ij}_t} + e^{-\frac{d^{ij}_t}{p_1}}\left(1- {d_1}/{d^{ij}_t} \right),\quad &\text{otherwise,}
\end{cases} 
\end{equation}
where $d_1 = \max\{460\log_{10}(h_j)-700,18\}$ and $p_1 = 4300\log_{10}(h_j)-3800$. If $h_j \in [100,300]$ m, we assume the LoS probability is 1.
For the path loss between BS $i$ and UAV $j$, the probabilistic mean path loss value is
\begin{flalign}
    L^{ij}_t = L_i^{\mathrm{LoS}}P_{\mathrm{LoS}}(r^{ij}_t) +L_i^{\mathrm{NLoS}}P_{\mathrm{NLoS}}(r^{ij}_t),
\end{flalign}
where  $L_i^{\mathrm{LoS}}$ and $L_i^{\mathrm{NLoS}}$ indicate the LoS and NLoS path loss, respectively.
Finally, the average received power from BS $i$ is
\begin{equation}
    P^{ij}_t = P_T + B^{ij}_t - L^{ij}_t,
\end{equation}
where $P_T$ is BS transmit power. The BS-UAV link is determined by the strength of the received signal $P^{ij}_t$, and the related signal-to-interference-plus-noise-ratio (SINR) is
\begin{align}
            \text{SINR}^{ij}_t = \frac{P^{ij}_t}{\sum_{i'=1, i' \neq i}^{N_R}P^{i'j}_t + \sigma_0^2},
\end{align}
where $\sigma_0^2$ is the additive white Gaussian noise power.

\subsection{UAV-HAPS Channel Model}

We assume each UAV carries a single antenna~\cite{ren2023handoff}, whereas the HAPS platform employs an array of multiple antennas. 
Then the HAPS can generates a large number of spot‐beams.
We fix the carrier frequency at 2\,GHz for every UAV–HAPS connection and assign each link to an orthogonal channel. Given the total available bandwidth \(B_{\max}\), we seek an optimal partitioning among these channels: For UAV \(j\), the fraction of bandwidth allocated to its link with the HAPS is denoted \(b^{Hj}\), subject to $\sum_{j \in \mathcal{M}} b^{Hj} \le 1$.
Similarly, power distribution must respect the UAVs’ transmission constraints. Let \(P_{\max}\) be the maximum transmit power for any UAV; we denote the fraction of this power allocated to UAV \(j\)'s HAPS link by \(p^{Hj}\) and $0\le p^{Hj}\le1,\forall\,j \in \mathcal{M}$.

The UAV–HAPS links are modeled as LoS, and their large-scale attenuation follows the free-space path-loss model. Accordingly, the instantaneous channel gain can be written as \cite{alsharoa2020improvement}:
\begin{equation}\label{eq:channel_gain}
    B_{t}^{Hj}
    = B  \biggl(\frac{c}{4\pi\,d_{t}^{Hj}\,f_{c}}\biggr)^{2}
    \bigl|h^{Hj}\bigr|^{2},
\end{equation}
where \(c\) denotes the speed of light, \(d_{t}^{Hj}\) is the UAV-HASP distance, and \(f_{c}\) is the carrier frequency. \(B\) represents the directional antenna gain, and \(h^{Hj}\) is the small-scale Rician‐fading coefficient accounting for the LoS component.

Therefore, the resulting transmission rate over the UAV–HAPS link is 
\begin{equation}\label{eq:rate}
    R_{t}^{Hj}
    = b^{Hj}B_{\max}
    \log_{2}\!\biggl(
      1 \;+\;
      \frac{p^{Hj}P_{\max}\,B_{t}^{Hj}}
           {b^{Hj}B_{\max}\,N_{0}}
    \biggr).
\end{equation}

\begin{figure}[t]
\includegraphics[width=1\linewidth]{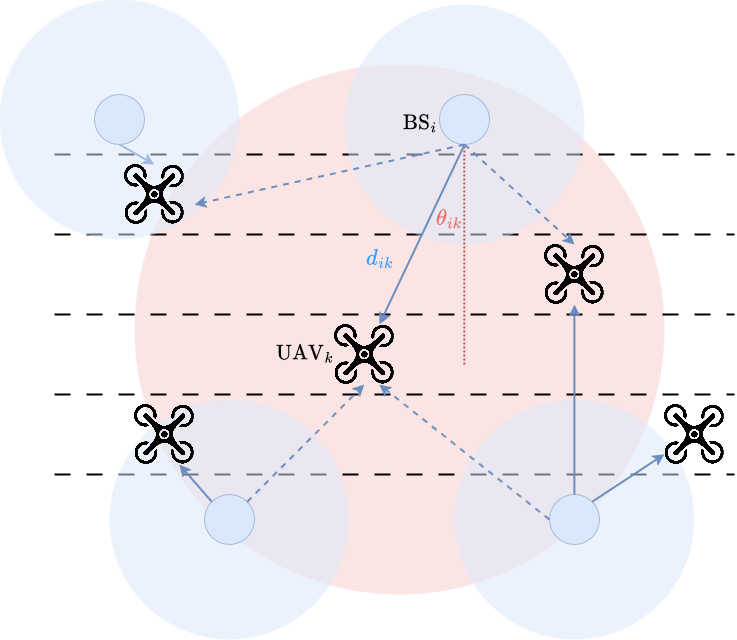}
\caption{Illustration of the proposed aerial network model (top view). Blue circles represent BSs; Solid/dash lines represent desired/interference link. 
}
\label{fig:BS_distributions_projection}
\end{figure}

\subsection{Weighted Data Rate with Handovers Constraints}

As UAVs navigate the aerial corridor, we distinguish two categories of handoffs: 1) Horizontal handoff: the UAV’s link is transferred between two terrestrial BSs; 2) Vertical handoff: the UAV’s connection shifts between terrestrial BSs and the HAPS.
Excessive handoff events can degrade the UAV’s achievable data rate, owing to the delays and potential failures associated with each transition. 
We assume each BS can access the possible UAV associations at each time instance, denoted by $n_i$. Then, based on the quota of each BS $i$, $Q_i \in [Q_R, Q_H]$, each UAV computes a \textit{weighted data rate metric} that encourages traffic load balancing and discourages unnecessary handoffs:
\begin{equation}
\label{eq:weight_data_rate}
    \text{WR}^{ij}_{t} = \frac{R^{ij}_{t}}{\min \left(Q_i, n_i \right)} (1 - \mu),
\end{equation}
where $\mu$ denotes the coefficient to discourage unnecessary handoffs. For instance, $\mu=0$ if keeping the previous BS, and $\mu=0.5$ if switching from terrestrial BS to HAPS and its traversal.

\section{Collaborative LLM Framework for Multi-UAV Control}
\label{sec:collab-llm}

As depicted in Fig.~\ref{fig:bdq_agent}, we employ a hierarchical two-level control architecture in which two collaborative LLM-based agents coordinate the UAV networks. The first LLM resides on the HAPS as a meta-controller overseeing global network operations. The second LLM is deployed at the network edge on each UAV, handling that individual UAV’s real-time motion decisions. The HAPS-level LLM continuously monitors the overall system state, including the total available HAPS bandwidth and each UAV’s instantaneous data rate, and issues high-level directives to the UAVs. In turn, the on-board UAV LLM controllers execute fine-grained flight control actions (e.g., speed and lane adjustments) in accordance with the meta-controller’s directives, while also responding to local environment conditions. This two-tier LLM framework enables cooperative decision-making across meta-control and edge-control levels, addressing both communication network performance and aerial traffic safety in a unified manner.

\subsection{HAPS Meta Decision Making Phase}

At the meta–control layer, a LLM deployed on the HAPS functions as a {global network orchestrator}.  Whenever the aggregate demand on the HAPS backhaul approaches its capacity, the LLM selects a subset of UAVs with the poorest HAPS link quality and instructs them to hand over to terrestrial BSs, thereby relieving congestion.  Once bandwidth becomes available, the meta‑controller authorizes the offloaded UAVs to re‑associate with the HAPS.  This policy maximizes the number of UAVs benefiting from the high‑throughput HAPS link while avoiding capacity violations and unnecessary handovers.

The decision process is modelled as an MDP 
$(\mathcal{S}_{\text{meta}},\mathcal{A}_{\text{meta}},\mathcal{P}_{\text{meta}},r_{\text{meta}})$.  
The state $\mathbf{s}_{\text{meta}}\in\mathcal{S}_{\text{meta}}$ captures the current HAPS load, per‑UAV data rates, and the availability of ground BS coverage.  
An action $\mathbf{a}_{\text{meta}}\in\mathcal{A}_{\text{meta}}$ specifies the set of UAVs to offload or recall.  
The transition kernel $\mathcal{P}_{\text{meta}}$ reflects the evolution of HAPS load and individual link qualities following these handovers.  
The reward of the HAPS is defined as
\begin{equation}
\label{eq:meta}
    r_{\text{meta}} = \underbrace{\eta_1\,\sum_{j \in \mathcal{M}}\text{WR}^{ij}_{t} - \eta_2\,\text{Sat}_{\text{HAPS}}}_{\text{throughput and saturation}} 
                 - \underbrace{\eta_3\,\mu}_{\text{handover penalty}},
\end{equation}
where the first term $\sum_{j \in \mathcal{M}}\text{WR}^{ij}_{t}$ weights the total throughput, the second term $\text{Sat}_{\text{HAPS}} \in \{0,1\}$ penalizes HAPS saturation events, and the third term $\eta_3\,\mu$ imposes a cost proportional to the number of enforced handovers. The LLM therefore learns a policy that maintains load balance, sustains high throughput, and minimizes disruptive switching.  
The overall procedure is summarized in Algorithm~\ref{alg:meta_controller}.

\begin{algorithm}[ht]
\caption{Meta-controller LLM for UAV Network Reconfiguration}
\label{alg:meta_controller}
\SetKwInOut{Input}{Input}
\SetKwInOut{Output}{Output}

\Input{Initial meta-state $\mathbf{s}_\text{meta}$, UAV set $\mathcal{U}$, HAPS bandwidth limit $C$, ground BS availability, threshold parameters}
\Output{Optimal UAV association actions $\mathbf{a}_\text{meta}$}
Initialize LLM parameters and policy network $\pi_\text{meta}$\;
\For{each timestep $t$}{
    Observe current meta-state $\mathbf{s}_t$ (UAV data rates, HAPS load, BS availability)\;\\
    Compute bandwidth usage: \\ $B_t \leftarrow \sum_{j \in \mathcal{J}} \text{WR}^{ij}_{t}$\;\\
    \eIf{$B_t > C$}{
        Identify UAV $j$ with lowest data rates or least critical tasks\;\\
        Action $\mathbf{a}_t \leftarrow$ Offload UAV $j$ to terrestrial BSs\;\\
    }{
        Identify offloaded UAV eligible for reattachment to HAPS\;\\
        Action $\mathbf{a}_t \leftarrow$ Reattach UAV $j$ to HAPS\;\\
    }
    Execute action $\mathbf{a}_t$\;\\
    UAVs update associations accordingly\;\\
    Observe next meta-state $\mathbf{s}_{t+1}$ and compute reward $r_{t}$ by Equation (\ref{eq:meta})
\\
    Update policy $\pi_\text{meta}$ using $(\mathbf{s}_t, \mathbf{a}_t, r_t, \mathbf{s}_{t+1})$\;\\
}
\Return learned optimal association policy $\pi^*_\text{meta}$\;
\end{algorithm}

\subsection{UAV Motion and Telecommunication Action Phase}

At the \textbf{edge-control} layer, each UAV embeds an onboard LLM responsible for real-time 3D trajectory and radio access decisions. In addition to longitudinal speed adjustments and lateral lane shifts, the UAV may alter its altitude whenever authorized by the HAPS meta-controller, yielding a unified horizontal–vertical maneuver set. Consequently, each UAV addresses its own Markov decision process (MDP), whose action space is influenced by the HAPS controller, while state transitions and rewards are observed locally.

\subsubsection{State Space}
Let $M_1$ denote the number of target UAVs within the decision horizon. Each UAV $j$ at time $t$ is characterized by
\[
\bigl\{x^j_t,\,y^j_t,\,z^j_t,\,v^j_t,\,\psi^j_t,\,n^{R,j}_t,\,n^{H,j}_t\bigr\},
\]
where $(x^j_t,y^j_t,z^j_t)$ is the 3D position, $v^j_t$ the forward velocity, $\psi^j_t$ the heading angle, and $n^{R,j}_t$ ($n^{H,j}_t$) the count of terrestrial BSs (HAPS channels) within service range that meet the target rate. Stacking these row vectors for all $j=1,\dots,M_1$ forms the global state
\[
\mathcal{S}_t =
\begin{bmatrix}
x^1_t & y^1_t & z^1_t & v^1_t & \psi^1_t & n^{R,1}_t & n^{H,1}_t \\[-1pt]
\vdots & \vdots & \vdots & \vdots & \vdots & \vdots & \vdots \\[-1pt]
x^{M_1}_t & y^{M_1}_t & z^{M_1}_t & v^{M_1}_t & \psi^{M_1}_t & n^{R,M_1}_t & n^{H,M_1}_t
\end{bmatrix}\!.
\]

\subsubsection{Two-Dimensional Action Space}\label{subsection:2daction}
At each decision instant, a \textbf{joint} action $(a_{\rm tele},a_{\rm tran})\in\mathcal{A}_{\rm tele}\times\mathcal{A}_{\rm tran}$ is selected. The set
\[
\mathcal{A}_{\rm tran}=\{a_{\rm tran}^1,\dots,a_{\rm tran}^5\}
\]
corresponds to \textbf{left-lane change}, \textbf{keep lane}, \textbf{right-lane change}, \textbf{accelerate}, and \textbf{decelerate}. Acceleration values follow the kinematic model in~\cite{yan2023multi}, so identical symbolic actions can produce different numeric rates.

The telecommunication choices
\[
\mathcal{A}_{\rm tele}=\{a_{\rm tele}^1,a_{\rm tele}^2,a_{\rm tele}^3\}
\]
denote BS selection strategies: $a_{\rm tele}^1$ selects the BS maximizing $\text{WR}_{ij}$ in~\eqref{eq:weight_data_rate}; $a_{\rm tele}^2$ uses the same metric with load threshold $\mu=0$, iteratively moving to the next candidate if saturated; and $a_{\rm tele}^3$ opts for the BS offering the highest instantaneous rate $R_{ij,t}$.

\subsubsection{Reward Design}
The transportation reward for UAV $j$ is
\begin{equation}
r^{\mathrm{j,tran}}_{t}=
w_1\!\Bigl(\frac{v^j_t-v_{\min}}{v_{\max}-v_{\min}}\Bigr)
- w_2\,\delta_c
- w_3\,\chi^j_t,
\end{equation}
where $\delta_c\in\{0,1\}$ indicates a collision and $\chi^j_t$ is the ratio of total lane changes to elapsed time. Negative returns are avoided to discourage premature termination.

For communications, the reward is
\begin{equation}
r^{\mathrm{j,tele}}_{t}=
w_4\,\text{WR}_{i^*,j,t}\bigl(1-\min(1,\xi^j_t)\bigr),
\end{equation}
where $\xi^j_t$ is the empirical handover probability up to time $t$. The weights $w_1,\dots,w_4$ prioritize safety ($w_2$) and connectivity ($w_4$) above secondary objectives.

Although the HAPS meta-controller operates on a slower timescale, its altitude offloading commands restrict each UAV’s vertical choices and, via $n^{H,j}_t$, influence the local state and rewards. This hierarchical LLM framework jointly optimizes network utilization and collision-free traffic flow, yielding a Pareto-efficient trade-off among throughput, handover overhead, and flight safety.

\section{Numerical Results and Discussions}

\begin{figure*}[t]
    \centering
    \begin{subfigure}{0.45\linewidth}
        \centering
        \includegraphics[width=\linewidth]{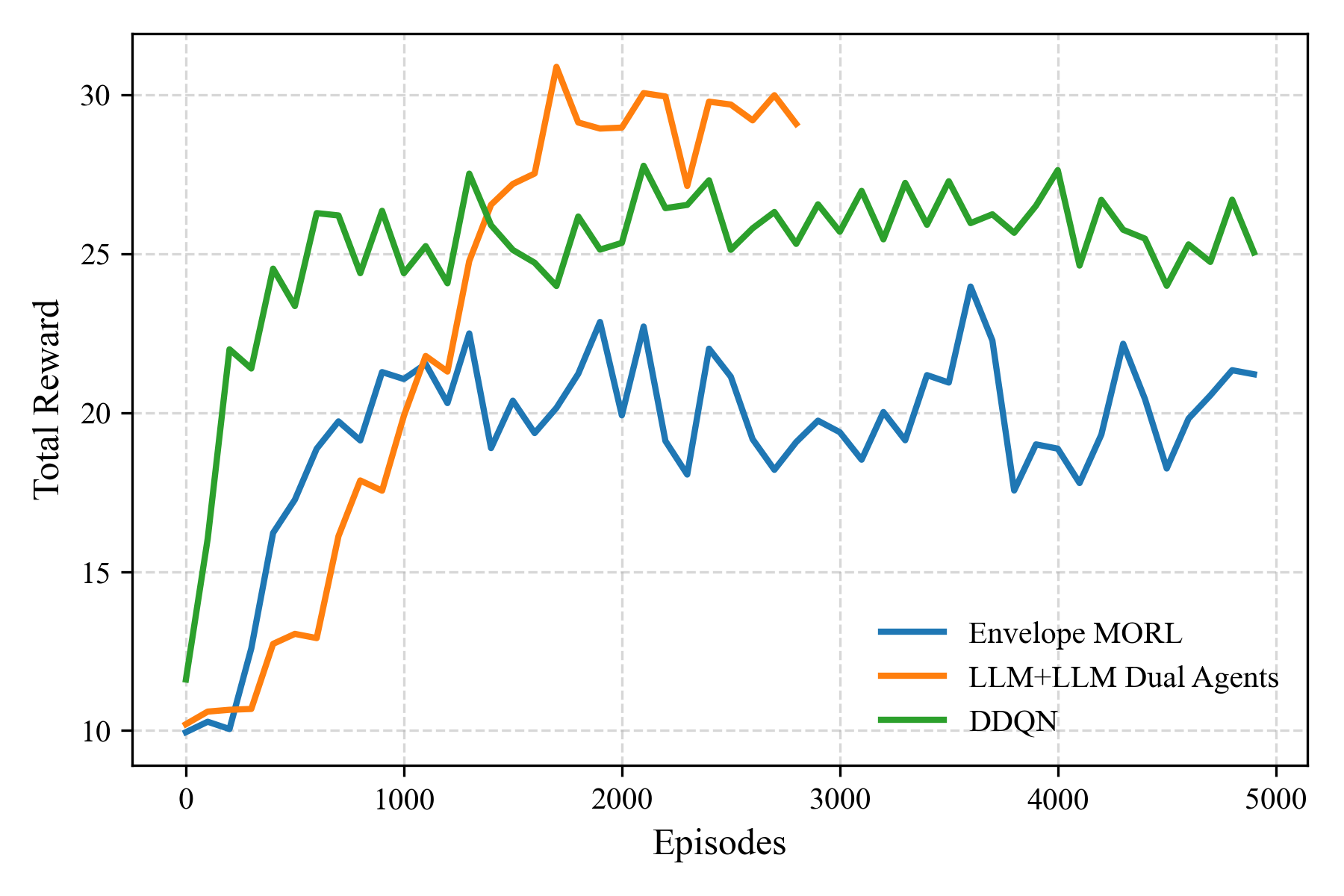}
        \caption{Total Reward}
        \label{fig:total_reward}
    \end{subfigure}
    \hfill
    \begin{subfigure}{0.45\linewidth}
        \centering
        \includegraphics[width=\linewidth]{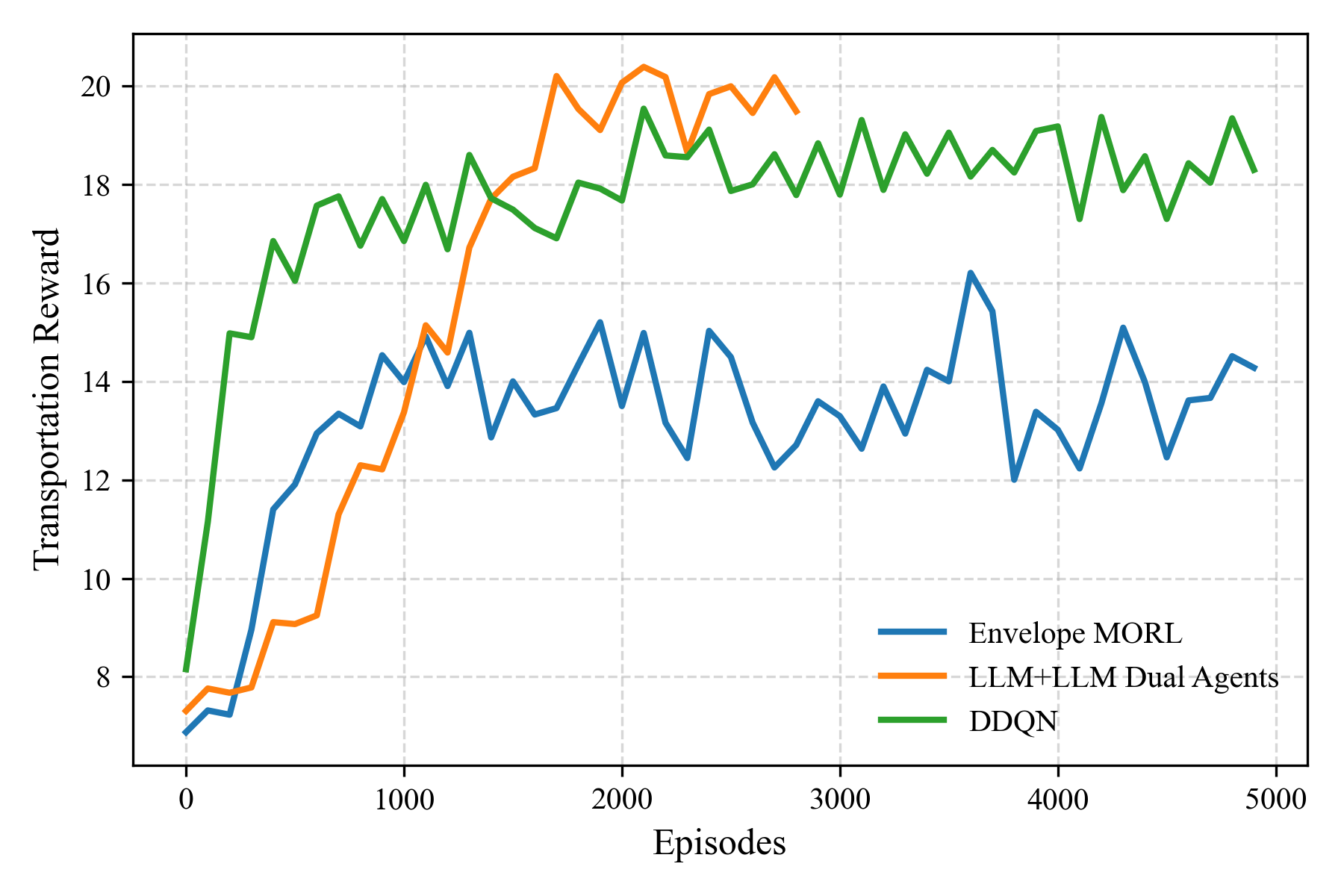}
        \caption{Transportation Reward}
        \label{fig:tran_reward}
    \end{subfigure}
    \vspace{1em}
    \begin{subfigure}{0.45\linewidth}
        \centering
        \includegraphics[width=\linewidth]{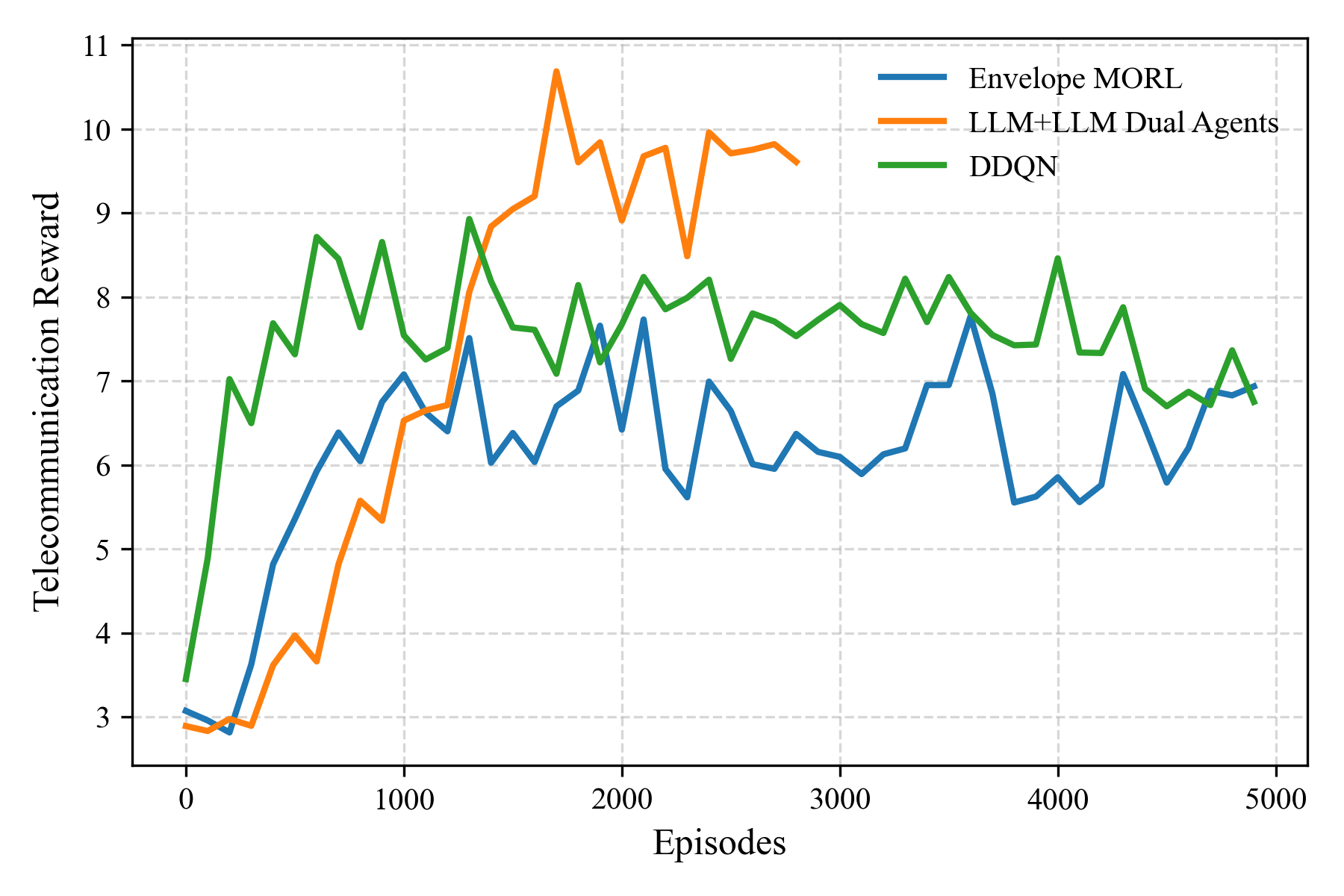}
        \caption{Telecommunication Reward}
        \label{fig:tele_reward}
    \end{subfigure}
    \hfill
    \begin{subfigure}{0.45\linewidth}
        \centering
        \includegraphics[width=\linewidth]{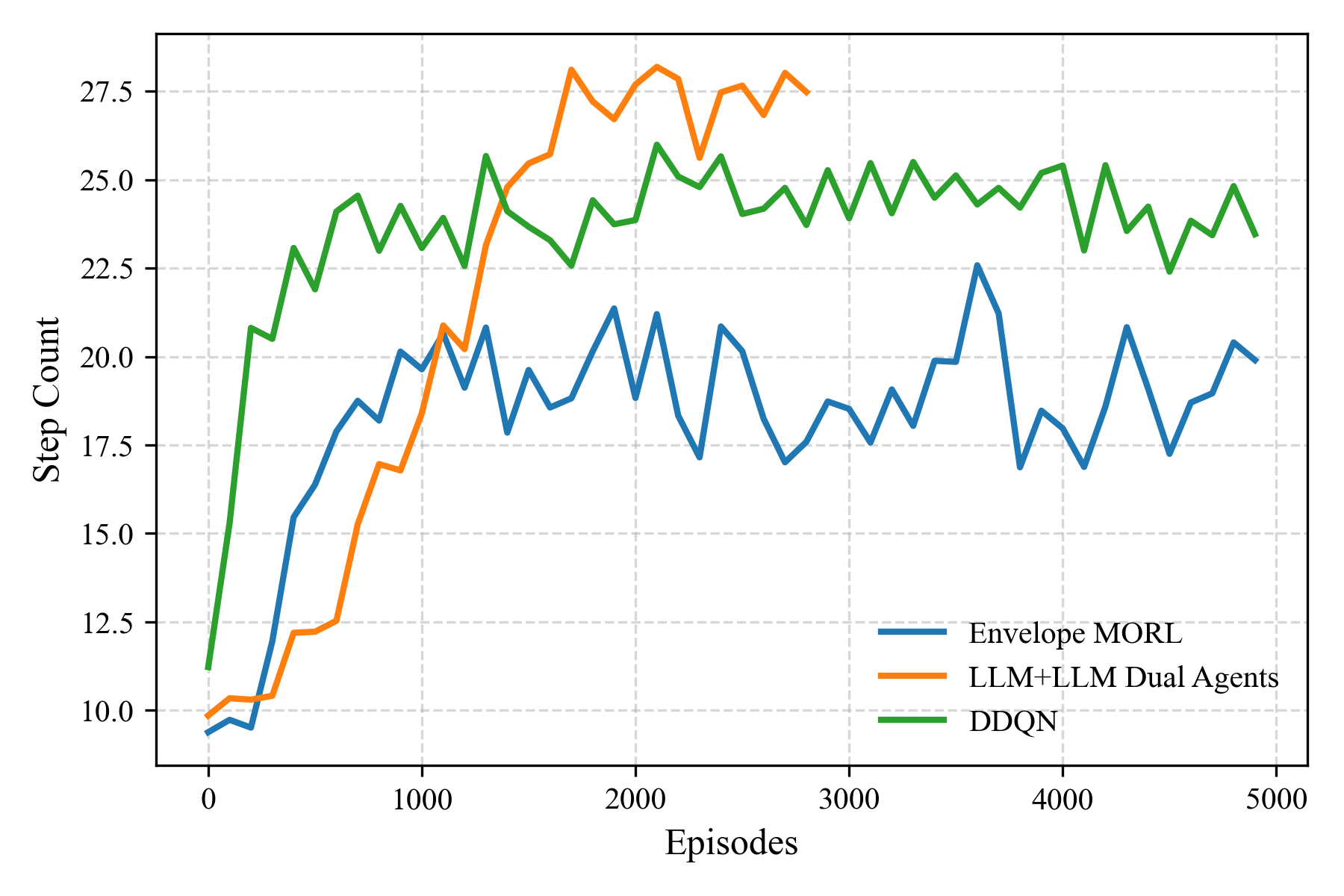}
        \caption{Step Count}
        \label{fig:step_count}
    \end{subfigure}
    \caption{Performance Comparison}
    \label{fig:multi_metric_comparison}
\end{figure*}


\begin{figure*}[t]
    \centering
    \begin{subfigure}{0.45\linewidth}
        \centering
        \includegraphics[width=\linewidth]{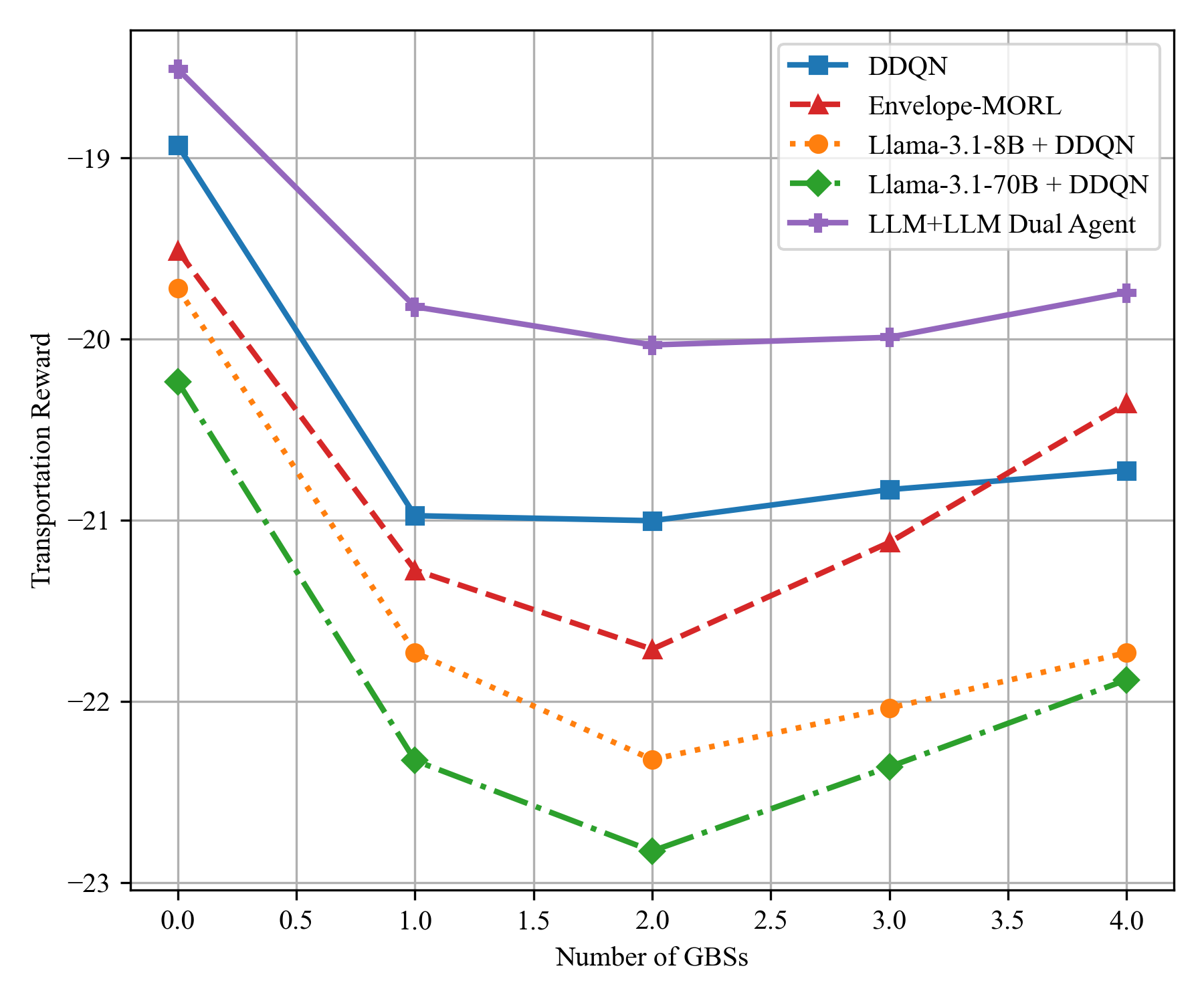}
        \caption{Transportation Cost}
        \label{fig:trans_cost}
    \end{subfigure}
    \hfill
    \begin{subfigure}{0.45\linewidth}
        \centering
        \includegraphics[width=\linewidth]{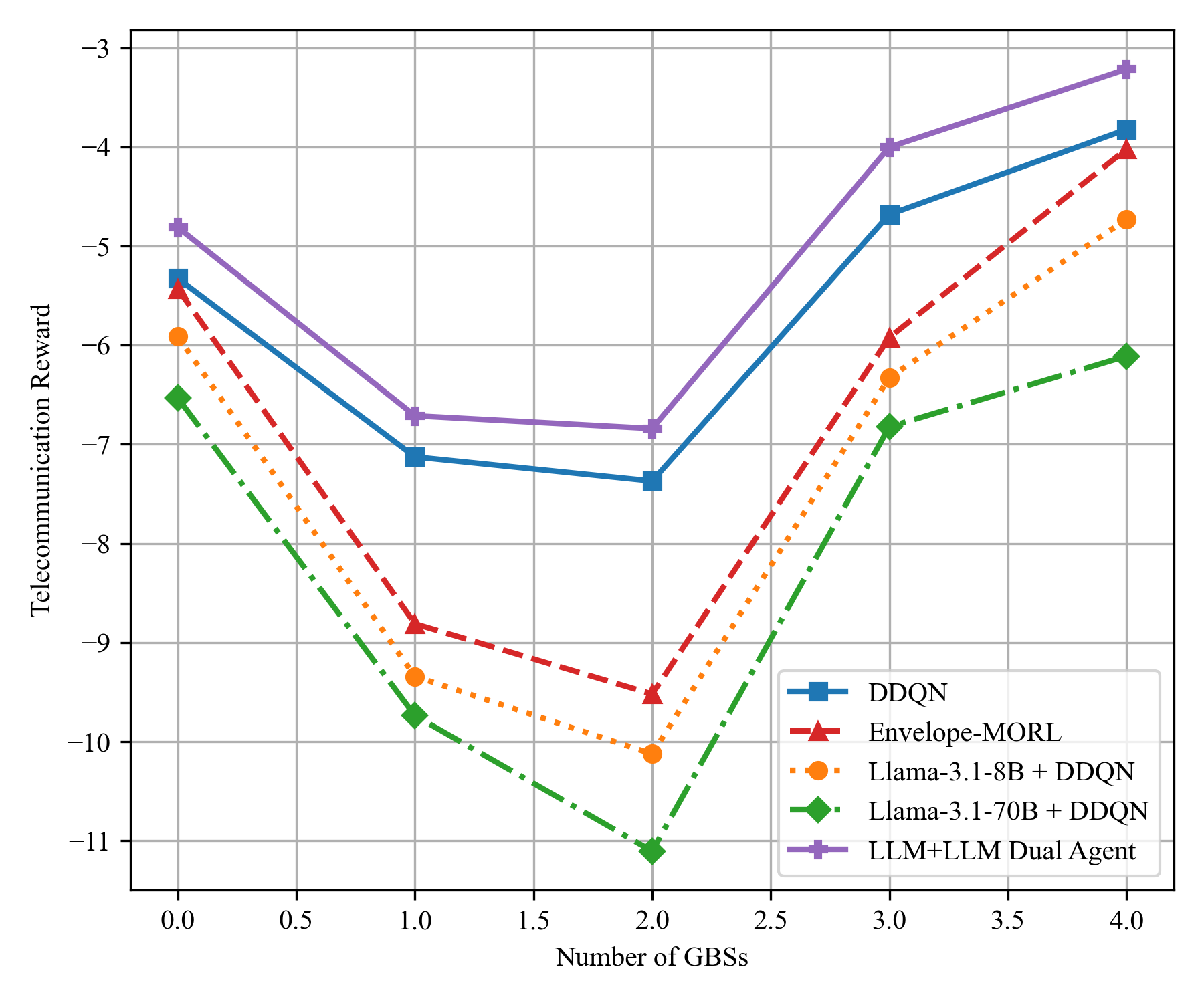}
        \caption{Telecommunication Cost}
        \label{fig:tele_cost}
    \end{subfigure}
    \vspace{1em}
    \begin{subfigure}{0.45\linewidth}
        \centering
        \includegraphics[width=\linewidth]{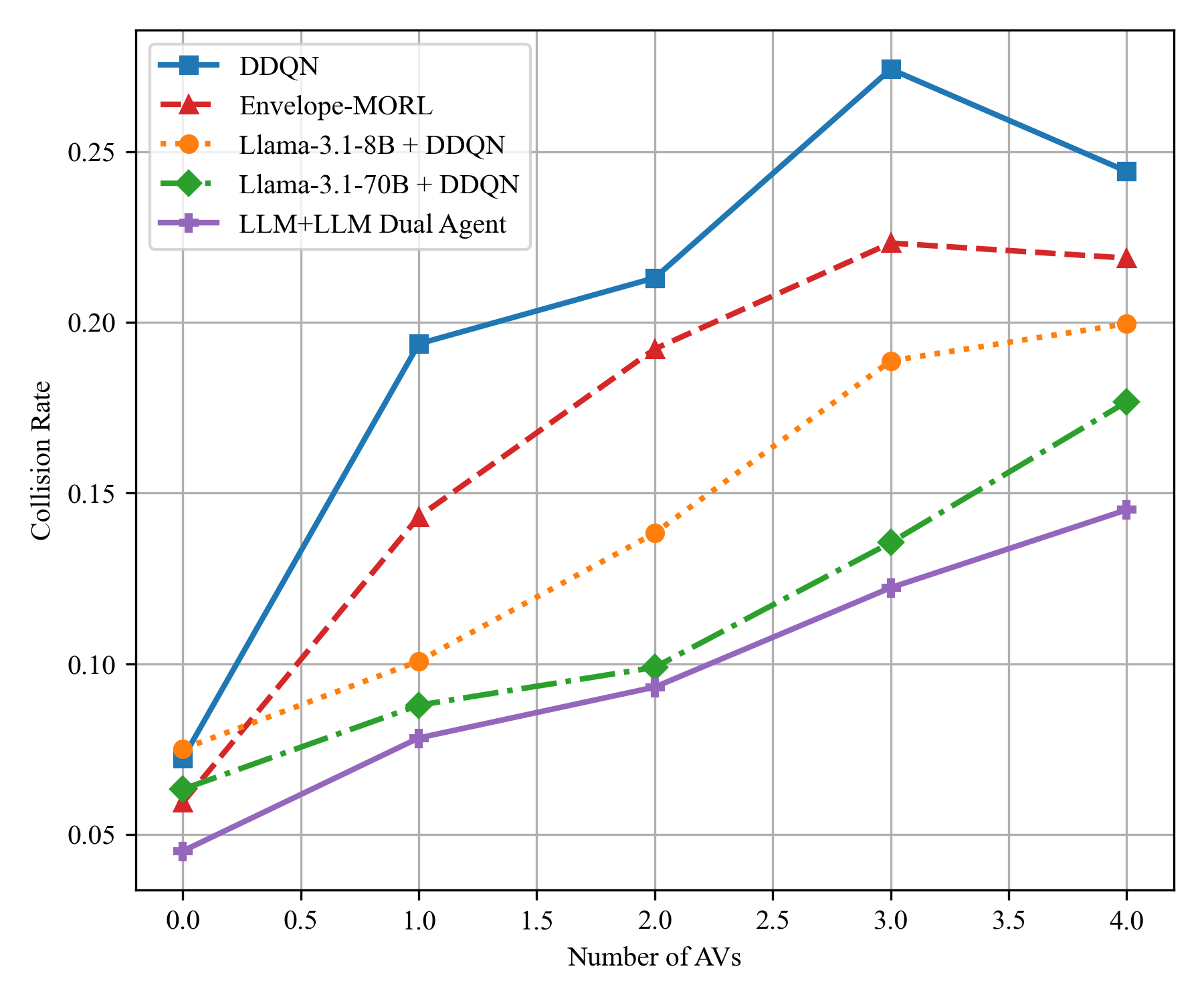}
        \caption{UAVs Collision Rate}
        \label{fig:collision_rate}
    \end{subfigure}
    \hfill
    \begin{subfigure}{0.45\linewidth}
        \centering
        \includegraphics[width=\linewidth]{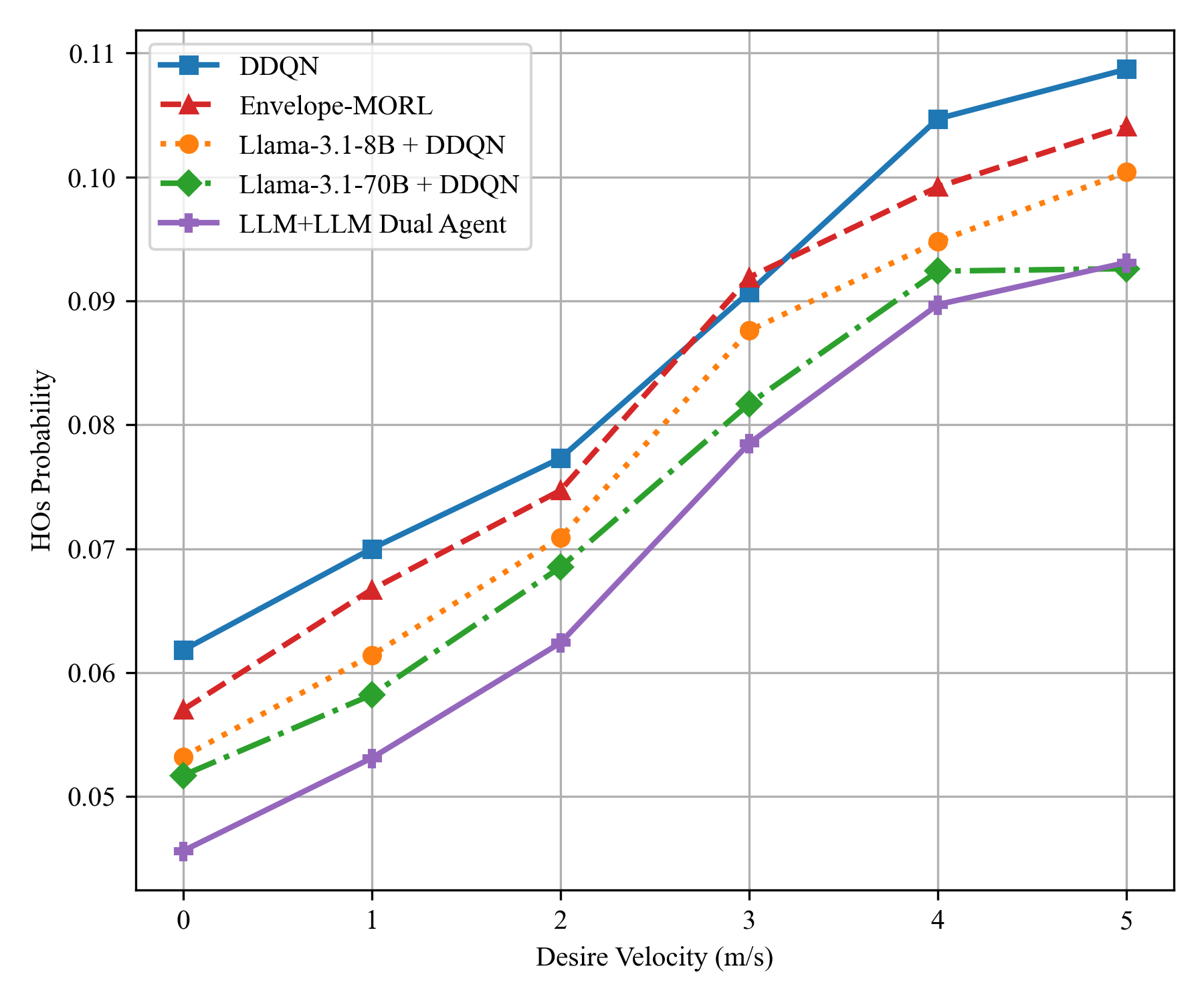}
        \caption{Step Count}
        \label{fig:eval_step_count}
    \end{subfigure}
    \caption{Evaluation Comparison}
    \label{fig:evaluation_comparison}
\end{figure*}

\subsection{Experiment Setup}
We consider a scenario with a total of five UAVs ($N_{\mathrm{UAV}} = 5$) navigating a five-lane aerial highway, with flying speeds ranging from 5~m/s to 20~m/s. GBSs operate at a carrier frequency of 2.1~GHz and transmit at a power of $P_T = 40$~dBm. Each BS supports a maximum of three UAV users and is deployed in varying quantities (5, 10, 15, or 20) to evaluate network scalability. 

Channel propagation is modeled using both LoS and NLoS path loss coefficients, set to $\eta_{\mathrm{LoS}} = 1$ and $\eta_{\mathrm{NLoS}} = 20$, respectively. The received signal power is constrained within the range $P_{\min} = -100$~dBm to $P_{\max} = -80$~dBm.

All LLM-driven simulations are conducted using the Ollama framework on a Linux host equipped with dual Intel E5-2650 v4 Broadwell CPUs (2.2~GHz) and two NVIDIA P100 GPUs, each with 16~GB of HBM2 memory. The maximum duration of each episode is capped at 30 time steps.

We benchmark our proposed framework against the following state-of-the-art deep reinforcement learning (DRL) and LLM-enhanced hybrid baselines: \\ 
\textbf{1) DDQN:} A buffer-aware Double Deep Q-Network that incorporates Lyapunov optimization to select bitrate levels under buffer occupancy constraints~\cite{yan2022reinforcement}.\\
\textbf{2) Envelope-MORL:} A generalized multi-objective reinforcement learning approach that leverages envelope updates to manage trade-offs among latency, throughput, and handover frequency in vehicular networks~\cite{yan2024generalized}.\\
\textbf{3) LLM+DDQN:} A hybrid system where a LLM offers semantically-informed guidance for autonomous driving decisions, while a DDQN agent concurrently optimizes V2I communication strategies~\cite{yan2025hybrid}.

\subsection{Training-Phase Convergence}

Fig.~\ref{fig:multi_metric_comparison} confirms that our \textit{LLM-LLM Dual Agents} outperform all benchmark methods across multiple metrics.

\textbf{(a)~Total Reward:} All methods exhibit steady improvements, but the Dual-Agent variant converges approximately an order of magnitude faster (within $\sim\!1.5\!\times\!10^{3}$ episodes) and peaks at a reward of $30$. In contrast, DDQN saturates near $23$, while Envelope-MHRL plateaus below $20$.

\textbf{(b)~Transportation Reward:} A similar performance ranking is observed. After convergence, the Dual-Agent model consistently achieves ${\sim}25\%$ higher transportation rewards than DDQN, indicating more assertive—yet safe—speed and lane-selection strategies.

\textbf{(c)~Telecommunication Reward:} The language-driven policies lead to enhanced V2I throughput. The Dual-Agent curve maintains a $2$–3 unit advantage over DDQN throughout most of the training horizon, while Envelope-MHRL fluctuates between the two baselines.

\textbf{(d)~Step Count:} All algorithms show an increasing trend in average episode length, suggesting smoother and collision-free trajectories. The Dual-Agent policy again reaches the maximum episode length the fastest, highlighting its efficient exploration and stable decision-making early in training.

\subsection{Evaluation under Varying Traffic Density}

Fig.~\ref{fig:evaluation_comparison} presents cost-oriented metrics averaged over the final 200 episodes as the number of UAVs increases from 5 to 40. Five schemes are evaluated: \textit{DDQN}, \textit{Envelope MHRL}, \textit{Llama 3.1 8B + DDQN}, \textit{Llama 3.1 70B + DDQN}, and the proposed \textit{LLM-LLM Dual Agent}.

\textbf{(a)~Total reward:}  
The reward initially decreases as the density increases from 5 to 20 UAVs, due to platooning effects, and then increases again as congestion begins to dominate. Throughout the range, the LLM-LLM framework consistently yields the highest total rewards, with an average improvement of approximately $16.3\%$ over the strongest baseline.

\textbf{(b)~Telecommunication reward:}  
A similar U-shaped pattern is observed. The LLM-LLM approach maintains the best performance, whereas DDQN suffers significantly when the number of UAVs exceeds 25, primarily due to uncoordinated and frequent handovers.

\textbf{(c)~Collision rate:}  
While all methods keep the collision rate below $0.12$ at low densities, it increases sharply when $M > 30$. The proposed dual LLM agent effectively limits the rate to below $0.08$, surpassing DDQN by up to $21\%$.

\textbf{(d)~Average steps:}  
The average number of steps grows nearly linearly with traffic density, reflecting increased queuing and decreased effective velocity. The dual agent system requires fewer additional steps compared to DDQN, demonstrating the robustness of language-guided exploration under congested conditions.

Overall, across both the training and evaluation phases, the proposed dual LLM agent framework achieves a clear Pareto improvement. It delivers higher cumulative rewards and reduced operational costs while maintaining lower collision rates in dense traffic. The purely language-driven control architecture consistently obtains the best episodic performance in both transportation and telecommunication domains.

\section{Conclusion and Future Work}

LLMs have demonstrated remarkable potential across various domains, and this work explores their capabilities in enhancing UAV mobility and communication control. We propose a novel hierarchical and collaborative decision-making framework that leverages LLMs to guide UAVs in dynamic multi-agent environments. The experimental results validate the superiority of our approach over several state-of-the-art baselines in terms of both transportation efficiency and telecommunication quality.

In future work, we aim to extend this framework by incorporating more accurate UAV localization and positioning mechanisms, especially in GPS-denied or urban canyon environments. We are also interested in integrating multi-modal sensory inputs and exploring online adaptation strategies to further improve the robustness and generalizability of LLM-driven decision-making in intelligent aerial transportation systems.


\bibliography{main.bib}

\begin{thebibliography}{21}
\providecommand{\natexlab}[1]{#1}
\providecommand{\url}[1]{\texttt{#1}}
\expandafter\ifx\csname urlstyle\endcsname\relax
  \providecommand{\doi}[1]{doi: #1}\else
  \providecommand{\doi}{doi: \begingroup \urlstyle{rm}\Url}\fi

\bibitem[3GPP(2018)]{3gpp777}
3GPP.
\newblock Study on enhanced {LTE} support for aerial vehicles (release 15), {TR} 36.777.
\newblock Jun. 2018.

\bibitem[Alsharoa \& Alouini(2020)Alsharoa and Alouini]{alsharoa2020improvement}
Alsharoa, A. and Alouini, M.-S.
\newblock Improvement of the global connectivity using integrated satellite-airborne-terrestrial networks with resource optimization.
\newblock \emph{IEEE Transactions on Wireless Communications}, 19\penalty0 (8):\penalty0 5088--5100, 2020.

\bibitem[Chen et~al.(2020)Chen, Lin, Khan, and Mozaffari]{chen2020efficient}
Chen, Y., Lin, X., Khan, T., and Mozaffari, M.
\newblock Efficient drone mobility support using reinforcement learning.
\newblock In \emph{Proc. IEEE Wireless Commun. Network. Conf. (WCNC)}, pp.\  1--6, 2020.

\bibitem[Cherif(2022)]{cherif2022cellular}
Cherif, N.
\newblock \emph{Cellular-connected {UAV} in Next-Generation Wireless Networks}.
\newblock PhD thesis, Universit{\'e} d'Ottawa/University of Ottawa, 2022.

\bibitem[Cherif et~al.(2021)Cherif, Jaafar, Yanikomeroglu, and Yongacoglu]{cherif20213d}
Cherif, N., Jaafar, W., Yanikomeroglu, H., and Yongacoglu, A.
\newblock 3d aerial highway: The key enabler of the retail industry transformation.
\newblock \emph{IEEE Communications Magazine}, 59\penalty0 (9):\penalty0 65--71, 2021.

\bibitem[Cherif et~al.(2023)Cherif, Jaafar, Yanikomeroglu, and Yongacoglu]{cherif2023rl}
Cherif, N., Jaafar, W., Yanikomeroglu, H., and Yongacoglu, A.
\newblock Rl-based cargo-uav trajectory planning and cell association for minimum handoffs, disconnectivity, and energy consumption.
\newblock \emph{IEEE Transactions on Vehicular Technology}, 2023.

\bibitem[Kotarski et~al.(2020)Kotarski, Piljek, and Kasa{\'c}]{kotarski2020design}
Kotarski, D., Piljek, P., and Kasa{\'c}, J.
\newblock Design considerations for autonomous cargo transportation multirotor uavs.
\newblock \emph{Self-Driving Vehicles and Enabling Technologies}, 2020.

\bibitem[Kurt et~al.(2021)Kurt, Khoshkholgh, Alfattani, Ibrahim, Darwish, Alam, Yanikomeroglu, and Yongacoglu]{kurt2021vision}
Kurt, G.~K., Khoshkholgh, M.~G., Alfattani, S., Ibrahim, A., Darwish, T.~S., Alam, M.~S., Yanikomeroglu, H., and Yongacoglu, A.
\newblock A vision and framework for the high altitude platform station (haps) networks of the future.
\newblock \emph{IEEE Communications Surveys \& Tutorials}, 23\penalty0 (2):\penalty0 729--779, 2021.

\bibitem[Li et~al.(2020)Li, Tan, Liu, Vijayakumar, Kumar, and Alazab]{li2020novel}
Li, X., Tan, J., Liu, A., Vijayakumar, P., Kumar, N., and Alazab, M.
\newblock A novel uav-enabled data collection scheme for intelligent transportation system through uav speed control.
\newblock \emph{IEEE Transactions on Intelligent Transportation Systems}, 22\penalty0 (4):\penalty0 2100--2110, 2020.

\bibitem[Qiu et~al.(2024)Qiu, Bakirtzis, Wassell, Song, Zhang, and Wang]{qiu2024large}
Qiu, K., Bakirtzis, S., Wassell, I., Song, H., Zhang, J., and Wang, K.
\newblock Large language model-based wireless network design.
\newblock \emph{IEEE Wireless Communications Letters}, 2024.

\bibitem[Ren et~al.(2023)Ren, Abbasi, Kurt, Yanikomeroglu, and Chen]{ren2023handoff}
Ren, Q., Abbasi, O., Kurt, G.~K., Yanikomeroglu, H., and Chen, J.
\newblock Handoff-aware distributed computing in high altitude platform station (haps)--assisted vehicular networks.
\newblock \emph{IEEE Transactions on Wireless Communications}, 2023.

\bibitem[Treiber \& Kesting(2013)Treiber and Kesting]{Treiber2013}
Treiber, M. and Kesting, A.
\newblock \emph{Traffic Flow Dynamics: Data Models and Simulation}.
\newblock Springer-Verlag Berlin Heidelberg, Sensors, 2013.

\bibitem[Xu et~al.(2024)Xu, Niyato, Kang, Xiong, Mao, Han, Kim, and Letaief]{xu2024large}
Xu, M., Niyato, D., Kang, J., Xiong, Z., Mao, S., Han, Z., Kim, D.~I., and Letaief, K.~B.
\newblock When large language model agents meet 6g networks: Perception, grounding, and alignment.
\newblock \emph{IEEE Wireless Communications}, 2024.

\bibitem[Yan \& Tabassum(2022)Yan and Tabassum]{yan2022reinforcement}
Yan, Z. and Tabassum, H.
\newblock Reinforcement learning for joint v2i network selection and autonomous driving policies.
\newblock In \emph{GLOBECOM 2022 - 2022 IEEE Global Communications Conference}, pp.\  1241--1246, 2022.
\newblock \doi{10.1109/GLOBECOM48099.2022.10001396}.

\bibitem[Yan \& Tabassum(2024)Yan and Tabassum]{yan2024generalized}
Yan, Z. and Tabassum, H.
\newblock Generalized multi-objective reinforcement learning with envelope updates in urllc-enabled vehicular networks.
\newblock \emph{arXiv preprint arXiv:2405.11331}, 2024.

\bibitem[Yan et~al.(2023)Yan, Jaafar, Selim, and Tabassum]{yan2023multi}
Yan, Z., Jaafar, W., Selim, B., and Tabassum, H.
\newblock Multi-uav speed control with collision avoidance and handover-aware cell association: Drl with action branching.
\newblock In \emph{GLOBECOM 2023-2023 IEEE Global Communications Conference}, pp.\  5067--5072. IEEE, 2023.

\bibitem[Yan et~al.(2025)Yan, Zhou, Tabassum, and Liu]{yan2025hybrid}
Yan, Z., Zhou, H., Tabassum, H., and Liu, X.
\newblock Hybrid llm-ddqn-based joint optimization of v2i communication and autonomous driving.
\newblock \emph{IEEE Wireless Communications Letters}, 14\penalty0 (4):\penalty0 1214--1218, 2025.
\newblock \doi{10.1109/LWC.2025.3539638}.

\bibitem[Yu et~al.(2022)Yu, Liu, Gao, Zhang, and Zhang]{yu2022deep}
Yu, J., Liu, X., Gao, Y., Zhang, C., and Zhang, W.
\newblock Deep learning for channel tracking in irs-assisted uav communication systems.
\newblock \emph{IEEE Transactions on Wireless Communications}, 21\penalty0 (9):\penalty0 7711--7722, 2022.

\bibitem[Zhang et~al.(2022)Zhang, Zong, Zhang, Dou, and Tian]{zhang2022game}
Zhang, R., Zong, Q., Zhang, X., Dou, L., and Tian, B.
\newblock Game of drones: Multi-uav pursuit-evasion game with online motion planning by deep reinforcement learning.
\newblock \emph{IEEE Transactions on Neural Networks and Learning Systems}, 2022.

\bibitem[Zhou et~al.(2024)Zhou, Hu, Yuan, Yuan, Wu, Liu, and Zhang]{zhou2024large}
Zhou, H., Hu, C., Yuan, D., Yuan, Y., Wu, D., Liu, X., and Zhang, C.
\newblock Large language model (llm)-enabled in-context learning for wireless network optimization: A case study of power control.
\newblock \emph{arXiv preprint arXiv:2408.00214}, 2024.

\bibitem[Zhou et~al.(2025)Zhou, Hu, Yuan, Yuan, Wu, Chen, Tabassum, and Liu]{zhou2025large}
Zhou, H., Hu, C., Yuan, D., Yuan, Y., Wu, D., Chen, X., Tabassum, H., and Liu, X.
\newblock Large language models for wireless networks: An overview from the prompt engineering perspective.
\newblock \emph{IEEE Wireless Communications}, 2025.

\end{thebibliography}
\bibliographystyle{icml2025}

\newpage
\appendix
\onecolumn



The following presents some supplementary experiments and discussions for “\textit{Hierarchical and Collaborative LLM-Based Control for Multi-UAV Motion and Communication in Integrated Terrestrial and Non-Terrestrial Networks}”.
In particular, we first present detailed example prompts that we used in the experiments to guide LLMs for UAV motion planning and communication decision-making. 
%
%
This document serves as supplementary material for readers to better understand how LLMs can be applied for UAV communication and motion planning dual optimization.

\section{HAPS Meta Action Selection}
Here is the example input on ChatGPT-4o for HAPS meta action Selection

\begin{mdframed}
\textbf{Task Description}: You are the \textbf{HAPS meta‑controller} overseeing UAV associations in an integrated terrestrial+HAPS network.  
Your job is to keep aggregate HAPS load under its capacity while maximising overall system throughput and minimising unnecessary handovers.

\textbf{Task Goal}:
\begin{itemize}
    \item Maintain total HAPS bandwidth usage $B_t$ \textbf{$\le C$} (capacity).
    \item Offload the least‑efficient UAVs when $B_t>C$; recall them when spare capacity exists.
    \item Minimise the number of enforced handovers and avoid service saturation.
\end{itemize}

\textbf{Environment Features}:
\begin{itemize}
    \item \textbf{UAV ID}: Index of the UAV under control.
    \item \textbf{Link}: Current serving station (\texttt{HAPS} or \texttt{TBS}).
    \item \textbf{Data Rate} [$\mathrm{Mbps}$]: Instantaneous throughput $R_{ij}^{t}$.
    \item \textbf{Task Priority}: Integer 1–5 (1 = mission‑critical, 5 = delay‑tolerant).
\end{itemize}
\noindent HAPS capacity: $C = 100\;\mathrm{Mbps}$.

\textbf{Current Meta‑state}:  
\[
\begin{array}{c|c|c|c}
\text{UAV ID} & \text{Link} & R_{ij}^{t}\;(\text{Mbps}) & \text{Priority}\\\hline
1 & \texttt{HAPS} & 28 & 2\\
2 & \texttt{HAPS} & 26 & 3\\
3 & \texttt{HAPS} & 24 & 1\\
4 & \texttt{HAPS} & 22 & 4\\
5 & \texttt{TBS}  & 18 & 3\\
\end{array}
\]
Total HAPS load $\;B_t = 28 + 26 + 24 + 22 = 100\;\mathrm{Mbps}$.

\textbf{Experience Replay}: The previous step was flagged as a \textbf{good} meta‑decision.

Here are some examples of \textbf{good} previous experiences (high reward):
\begin{itemize}
    \item \textbf{State}: $B_t=98$, UAV‐4 offloaded $\rightarrow$ \textbf{Action}: \texttt{Offload\{4\}} $\rightarrow$ \textbf{Reward}: $+1.20$
    \item \textbf{State}: $B_t=85$, UAV‐5 recalled  $\rightarrow$ \textbf{Action}: \texttt{Recall\{5\}}  $\rightarrow$ \textbf{Reward}: $+0.95$
\end{itemize}

Here are some examples of \textbf{poor} previous experiences (negative reward):
\begin{itemize}
    \item \textbf{State}: $B_t=112$, no action taken $\rightarrow$ \textbf{Action}: \texttt{Idle} $\rightarrow$ \textbf{Reward}: $-1.50$
\end{itemize}

\textbf{Rules}:  
Choose \textbf{one} of the following meta‑actions:
\begin{itemize}
    \item \texttt{Offload\{ID$_1$,ID$_2$,\dots\}} – detach the listed UAVs from HAPS to TBSs.
    \item \texttt{Recall\{ID$_1$,ID$_2$,\dots\}}  – move the listed UAVs back to HAPS.
    \item \texttt{Idle} – keep all current associations.
\end{itemize}

\textbf{Decision}: Provide the chosen action wrapped in \verb|<meta_action>|\,\verb|</meta_action>| and briefly justify your choice.
\end{mdframed}

LLM agent respond as

\begin{mdframed}
HAPS load is exactly at the limit ($B_t=100\;\mathrm{Mbps} = C$).  
A single network fluctuation could push us beyond capacity, so pre‑emptive relief is prudent.  
UAV‑4 has the \textbf{lowest data rate (22 Mbps)} and \textbf{lowest priority (4)} among HAPS‑attached nodes, making it the best candidate for offloading.  
No handover penalty will be incurred for UAV‑5 because it is already on a terrestrial BS.

\verb|<meta_action>Offload{4}</meta_action>|

\end{mdframed}

\section{UAV Motion and Communication Dual Actions Selection}

\begin{mdframed}
\textbf{Task Description}: You are an agent that \textbf{simultaneously} decides \textbf{(i)} the ego‑UAV’s driving maneuver on a multi‑lane highway and \textbf{(ii)} which base‑station policy (t1/t2/t3) to follow for V2I communication.

\textbf{Task Goal}:
\begin{itemize}
    \item \textbf{Transport}: Maximise velocity while avoiding collisions and reducing unnecessary lane changes.
    \item \textbf{Telecom}: Maximise weighted data‑rate, minimise handovers (HOs), and keep load balanced across BSs.
\end{itemize}

\textbf{Environment Features} (per UAV row):
\begin{itemize}
    \item \textbf{'x'} – Horizontal offset (m).
    \item \textbf{'y'} – Vertical offset (lane index).
    \item \textbf{'vx'} – Longitudinal speed (m/s).
    \item \textbf{'vy'} – Lateral speed (m/s); non‑zero means a lane change.
\end{itemize}
Additional telecom counters for the ego UAV: \texttt{gbs\_cnt}, \texttt{haps\_cnt}.  
Normalised bounds: $[100,100,20,20]$ for $x,y,vx,vy$.

\textbf{Observations}: Current discretised state for \textbf{three} UAVs (row 1 = ego):
\[
\underbrace{\begin{bmatrix}
2 & 4 & 3 & 1\\
3 & 1 & 0 & 3\\
4 & 5 & 2 & 2
\end{bmatrix}}_{\text{transport matrix}}
\qquad
\bigl[\texttt{gbs\_cnt}=2,\; \texttt{haps\_cnt}=1\bigr]
\]

\textbf{Experience Replay}: Last step was good.  

\textit{Good examples (top‑3):}
\begin{itemize}
  \item \textbf{State}: $[2,4,3,1,\,3,1,0,3,\,4,5,2,2,\,2,1]$, \textbf{Action}: \{\texttt{FASTER},\;\texttt{t1}\}, \textbf{Reward}: $+1.05$
  \item \textbf{State}: $[3,4,3,1,\dots]$, \textbf{Action}: \{\texttt{IDLE},\;\texttt{t1}\}, \textbf{Reward}: $+0.92$
\end{itemize}

\textit{Bad examples (top‑1):}
\begin{itemize}
  \item \textbf{State}: $[5,6,6,2,\dots]$, \textbf{Action}: \{\texttt{LANE\_LEFT},\;\texttt{t3}\}, \textbf{Reward}: $-1.20$
\end{itemize}

\textbf{Rules}:
\begin{itemize}
  \item Transport action $\in\{$\texttt{FASTER}, \texttt{SLOWER}, \texttt{LANE\_RIGHT}, \texttt{LANE\_LEFT}, \texttt{IDLE}$\}$.
  \item Telecom action $\in\{\texttt{t1},\texttt{t2},\texttt{t3}\}$.
  \item Output must contain \verb|<tran_action>|...\verb|</tran_action>| and \verb|<tele_action>|...\verb|</tele_action>|.
\end{itemize}

\textbf{Decision}: Provide both actions and justify briefly.
\end{mdframed}

The prompt is organised into six intuitive blocks from the viewpoint of a \textbf{UAV decision agent}:

\begin{enumerate}
    \item \textbf{Task Description} declares the overarching mission: maintain high–speed, collision‑free flight along the aerial highway.
    \item \textbf{Task Goal} – details the three optimisation targets for each UAV: (i) maximise forward velocity, (ii) avoid collisions, and (iii) minimise unnecessary lateral manoeuvres.
    \item \textbf{State Definition} – enumerates all environmental features inspected by the agent, namely the 3‑D position, velocity, heading, and the counts of viable \textbf{BS}/\textbf{HAPS} links.
    \item \textbf{Observations} – present a discretised post‑processing matrix of those features for the ego UAV and surrounding aircraft.
    \item \textbf{Experience Memory} – supplies two reference tables (``good’’ vs.\ ``bad’’ cases) that offer precedents to guide the LLM when it encounters unseen states.
    \item \textbf{Reply Rules} – constrain the model’s output to exactly one transport action and one telecom policy selected from the predefined action set, ensuring decisions remain focused and comparable across time‑steps.
\end{enumerate}

This structured natural‑language template encapsulates the entire optimisation problem while remaining fully compatible with the meta‑controller logic in Algorithm~1.

LLM agent respond is
\begin{mdframed}
The ego UAV is already centred in its lane and moving at a moderate speed (\texttt{vx}=3).  
Vehicle 2 ahead is slower, but there is ample gap on its left; a controlled acceleration keeps throughput high without risking collision.

For telecom, the load is light (\texttt{gbs\_cnt}=2, \texttt{haps\_cnt}=1) and recent good experiences indicate \texttt{t1} yields higher long‑term reward when HO penalties matter.

\verb|<tran_action>FASTER</tran_action>|

\verb|<tele_action>t1</tele_action>|
\end{mdframed}
\begin{itemize}
    \item \textbf{Edge deployment via \textsc{Ollama}.}  
    The LLM is hosted on nearby edge servers using the \textsc{Ollama} framework.  
    Locating inference close to the autonomous vehicles eliminates wide‑area round‑trip delays, keeping response times within the strict real‑time bounds required for autonomous driving.

    \item \textbf{Targeted experience prompts.}  
    To curb computational load, the agent supplies the LLM with only the five \textbf{most similar} \textbf{good} and \textbf{bad} experiences, determined by Euclidean distance in state space, rather than the entire replay buffer.  
    This focused context preserves decision quality while markedly accelerating inference.
\end{itemize}

\end{document}